# Nearest Prime Simplicial Complex for Object Recognition

*Junping Zhang[1], Ziyu Xie[1], and ** Stan Z. Li[2]

[1] Shanghai Key Laboratory of Intelligent Information Processing and School of Computer Science, Fudan University, Shanghai, China
[2] Institute of Automation, Chinese Academy of Sciences, Beijing, China.

**Abstract.** The structure representation of data distribution plays an important role in understanding the underlying mechanism of generating data. In this paper, we propose nearest prime simplicial complex approaches (NSC) by utilizing persistent homology to capture such structures. Assuming that each class is represented with a prime simplicial complex, we classify unlabeled samples based on the nearest projection distances from the samples to the simplicial complexes. We also extend the extrapolation ability of these complexes with a projection constraint term. Experiments in simulated and practical datasets indicate that compared with several published algorithms, the proposed NSC approaches achieve promising performance without losing the structure representation.

## 1 Introduction

The structure representation is important to understanding the underlying mechanism of generating data. To capture such structures, manifold learning algorithms [1,2] assume that data are generated from an underlying low-dimensional manifold. However, it is not easy to discover and preserve the topological structure hidden in the manifold. For example, [3] pointed out that the view and style-independent action manifolds, which are used to describe human activities, can be assumed to lie in a torus.

Persistent homology can effectively discover the topological invariants such as holes, which cannot be easily available by other means such as manifold learning algorithms [4]. The method first incrementally constructs nested families of simplicial complexes from point cloud data (PCD), and then computes the lifecycle of each possible topological invariant by placing the complexes within an evolutionary growth process. Finally, it extracts those truly topological invariants or features with longer lifecycle and removes topological noises [5]. However, how to employ it for practical applications (e.g., object recognition) remains unsolvable.

---
* Junping Zhang, Ziyu Xie, Email: jpzhang@fudan.edu.cn, ziyu.ryan@gmail.com
** Stan Z. Li, Email:szli@nlpr.ia.ac.cn

In this paper, we propose a novel method, called nearest prime simplicial complex approaches (NSC), to obtain a structure-preserving representation and achieve higher performance in object recognition. Specifically, we generate a nested family of simplicial complexes per class, and estimate a prime simplicial complex per class by weighting the lifecycles of alive topological structures. Then we classify objects based on the nearest projection distances from each object to simplices in these simplicial complexes. Furthermore, we also utilize a projection constraint term to enhance the extrapolation ability of NSC and prevent incorrect projection. The main contribution is that we extend the geometrical framework of simplicial complex to object recognition. Specifically, we propose NSC approaches to object recognition and show how to use them for point classification. Experiments in several simulated and practical datasets show that without losing the structure representation, the proposed NSC approaches attain promising performance compared with several well-known algorithms.

The remainder of this paper is organized as follows. In Section 2, we will introduce some preliminary of simplicial complex and give a brief survey on persistent homology. In Section 3 we will detail our proposed NSC algorithm. We evaluate the performance of the proposed NSC approaches in Section 4. We conclude the paper in Section 5.

## 2   Preliminary and Related work

In this section, we will introduce some preliminary of simplicial complex and the construction of simplicial complex, and survey the development of persistent homology.

### 2.1   Preliminary

The simplicial complex is a collection of simplices subject to some rules. The simplex and simplicial complex are defined as follows:

**Definition 1:** Let $\{v_0, v_1, \cdots, v_p\}$ be a geometrically independent set in $\mathbb{R}^N$. We define the $p$-simplex $\sigma$ spanned by $v_0, v_1, \cdots, v_p$ to be the set of all points $x$ of $\mathbb{R}^N$ such that [6]: $x = \sum_{i=0}^{p} \lambda_i v_i$, where $\sum_{i=0}^{p} \lambda_i = 1$ and $\forall i, \lambda_i \geqslant 0$.

In general, each $p$-simplex $\sigma$ has $p+1$ faces which are $(p-1)$-simplices. The face is obtained by deleting one of the vertices $v_0, v_1, \cdots, v_p$.

**Definition 2:** A simplicial complex $K$ in $\mathbb{R}^N$ is a collection of simplices in $\mathbb{R}^N$ such that [6]: 1) Every face of a simplex of $K$ is in $K$. 2) The intersection of any two simplices of $K$ is a face of each of them.

An illustration on the distinction between simplicial complex and non-simplicial complex is shown in Fig. 1. Obviously, the non-simplicial complex violates the second rule.

In this paper, we mainly utilize Lazywitness complexes, which behaves like Delaunay triangulations computed in the intrinsic geometry of the data set $\boldsymbol{X}$, to construct the simplicial complex for PCD.

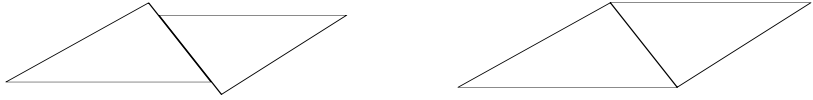

**Fig. 1.** The distinction between non-simplicial complex (left) and simplicial complex (right).

Specifically, we select a subset $\boldsymbol{Z} = \{v_1, \cdots, v_p\} \subset \boldsymbol{X}$ as the vertex set by using sampling techniques such as max-min sampling or random sampling methods at first. The max-min sampling method randomly extracts one point as the first vertex, and iteratively selects the next $n-1$ points that maximize the minimal distances between them and the previous vertices. With this way, the method generates a vertex set uniformly distributed around the data structure [4]. Then we utilize the remaining points as the witness points set $\{w_1, w_2, \cdots, w_q\}$ to determine which simplices occur in the complex [4].

More formally, let $\boldsymbol{D}$ be a $p \times q$ distance matrix, where $q$ denotes the number of the remaining points. Each element $\boldsymbol{D}(i,j)$ measures the distance between landmark point $i$ and witness point $j$. To discover a persistent topological invariant from PCD, we construct a nested family of simiplicial complexes $W(\boldsymbol{D}; R, f)$. Here $R$ is the radius of metric ball, and $f$ is a non-negative integer. If $f=0$, we define $m_j = 0, \quad \forall j = 1, 2, \cdots, q$. Otherwise, let $m_j$ be the $f$-th smallest entry of the $j$-th column of matrix $\boldsymbol{D}$. Then we utilize two rules to determine which simplices can be added into the complex [4]: 1) a 1-simplex $\sigma = [v_a v_b]$ will be added to $W(\boldsymbol{D}; R, f)$ iff there exists a witness $w_j$ $(1 \leqslant j \leqslant q)$ satisfying $\max(\boldsymbol{D}(a,j), \boldsymbol{D}(b,j)) \leq R + m_j$. 2) a $k$-simplex $[v_{a_0} v_{a_1} \cdots v_{a_k}]$ will be added to $W(\boldsymbol{D}; R, f)$ iff there exists a witness $w_j$ $(1 \leqslant j \leqslant q)$ satisfying $\max(\boldsymbol{D}(a_0, j), \boldsymbol{D}(a_1, j), \cdots, \boldsymbol{D}(a_k, j)) \leq R + m_j$.

Note that when the number of training samples is small, which is very common in object recognition domain, we can instead use the Rips complex method to obtain the nested families of simplicial complexes. Assuming that each point is a center of a closed Euclidean ball with radius $R$, Rips method iteratively builds a complex by forming a line in any two points if the balls of them are intersected [4].

## 2.2 Related work

Persistent homology is to discover some stable topological invariants from PCD. To achieve the goal, there are three crucial steps [5]: 1) selecting a subset that expresses the non-trivial topological attributes measured by homology groups, 2) measuring the importance of these subsets and 3) eliminating those topological attributes with the minimum number of side-effects.

[4,7] investigated the influence of sampling technique to the estimation of topological invariants. With persistent homology and sampling strategy, [4,7] discovered that image patches with edges lie in a Klein-bottle-shape space. [8] proposed to use geodesic Delaunay triangulation to reduce the number of samples, which is required to capture the topology of PCD.

To discover the topological structure from data cloud points, it is necessary to construct the simplicial complexes. There are several different complexes including Čech, Rips, Explicit, Witness and Lazywitness complexes in literatures. Let PCD be $X$, and the radius of PCD be $R$. Specifically, the Čech complex Čech$(X, R)$ means the nerve of the collection of metric balls $\{B(x_j), R/2\}$, $x_j \in X$, $j = 0, 1, \cdots, p$ [9], with vertex set $X$. [5] proposed $\alpha$-complexes, and estimated the invariants through computing the persistent Betti numbers. For saving storage space, Rips$(X, R)$ only stores the edges and vertices, and forms the largest simplicial complex that has the same 1-skeleton (i.e. vertices and edges) as Čech$(X, R)$. However, both of the two methods produce a very large amount of complexes, especially for large-scale PCD. To refine the efficiency, [4] proposed witness complex by selecting a group of landmark points and utilizing the remaining points as witness of the existence of simplicial complexes.

[10] employed Barcode technique to measure the importance of topological attributes. Furthermore, [11] applied the persistent homology to extract some topological features from character-shape point cloud data. [12] studied the smallest coverage issue in sensor networks based on the persistent homology. Assuming that stratified spaces consist of multiple manifolds or non-manifolds, each of which has varying dimension, [13] generalized the computation of persistent homology to that of intersection homology for better analyzing the stratified spaces. Moreover, [14] clustered data points into different stratified space using methods derived from kernel and cokernel persistent homology. [15] investigated the persistent homology of random fields and manifold learning. A major difficulty is that it is not easy to fill the gap between the persistent homology and practical applications.

## 3  Nearest Prime Simplicial Complex

In this section, we will detail the NSC approaches by dividing them into two parts.

### 3.1  Selecting prime simplicial complexes

To utilize the persistent homology for recognition, we propose three crucial steps including eliminating the redundant simplices, recording a recognition-related Barcode and selecting the prime simplicial complexes.

With the methods mentioned above, specifically, we can construct a filtered simplicial complex from the point cloud data by increasing $R$ from 0 to $\infty$. The filtered complex is an increasing sequence of simplicial complexes which determine an inductive system of homology groups [10]. To our research, we discover that in this sequence, a proper complex, named prime simplicial complex, is useful for recognition. The prime simplicial complex is a relatively stable complex from which we can capture the homology of the data's topological structure. For better understanding, an example is shown in Fig. 2.

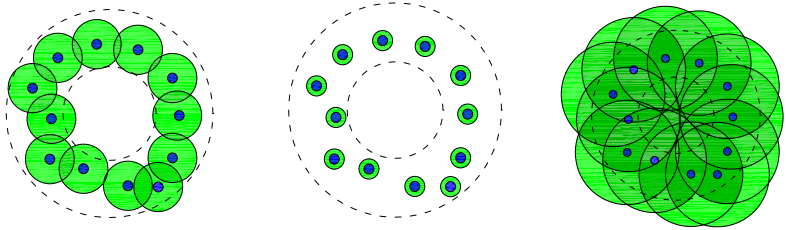

**Fig. 2.** We can construct a simplicial complex through metric balls with a radius $R$. A good choice of $R$ (left) induces a prime simplicial complex which can help us to capture the homology of an annulus from the union of balls. Meanwhile, the union of balls with incorrect radius will induce an incorrect structure representation (middle and right).

A $k$-simplex which is not a face of any $k+1$-simplices of the same complex is a relatively highest-dimensional simplex compared with its lower-dimensional ones. Note that we always focus on the relatively highest-dimensional simplices of the prime simplical complex since their faces which are lower-dimensional ones have been implicitly considered in our NSC approaches. For avoiding the repeated computation, we propose to remove these faces when constructing the prime simplicial complex. Here we give a pseudo-code based on Lazywitness complexes in Algorithm 1. Note that in line 3, the matrix $\boldsymbol{E}$ is calculated as:

$$\boldsymbol{E}(i,j) = \min_k \max(\boldsymbol{D}(i,k), \boldsymbol{D}^*(k,j)) - m_k \tag{1}$$

and in line 9, the lower-dimensional simplices will be removed after the merging procedure is completed.

Once the prime simplicial complex is constructed, we use Barcode technique to record the lifetime of each simplex belonging to the complex as the parameter $R$ increases until $R_{\max}$ is reached. We only consider the simplices which are still alive when $R = R_{\max}$. An example is shown in Fig. 3 [3].

Obviously, it is hard to find the best prime simplicial complex from the sequence. Therefore, we propose to select a radius $R^*$ based on the weighted lifecycles:

$$R^* = \frac{\sum_{i=1}^m \ell_i M_i}{\sum_{i=1}^m \ell_i} \tag{2}$$

---

[3] Note that our Barcode is different from that in [15]. The reason is that although [15]'s Barcode technique is a good way to describe the persistent homology by recording the birth and death time of some topological invariants, only the alive simplices are useful for our proposed NSC algorithm.

---

**Algorithm 1** Construct the Prime Simplicial Complex using Lazywitness Complexes

---

**input** Point Data $\boldsymbol{P}$, $R$, the ratio $r$, the family $f$
**output** the vertices of each simplex constructing the simplicial complex
1: Choose $p$ landmark points and $q$ witness points, where $p = \texttt{size}(P)/(r+1)$ and $q = p \cdot r$.
2: Compute the $p \times q$ matrix $\boldsymbol{D}$ of distances.
3: Compute the $p \times p$ matrix $\boldsymbol{E}$ with off-diagonal entries $\boldsymbol{E}(i,j) = R_{[v_i v_j]}$ which record the time when edge $v_i v_j$ appears.
4: Consider every two pairs $(i,j)$ where $i < j \leqslant p$
5: **if** $\boldsymbol{E}(i,j) \leqslant R$ **then**
6:     Add $[v_i v_j]$ to $W(\boldsymbol{D}; R, f)$.
7:     Remove $[v_i], [v_j]$ from $W(\boldsymbol{D}; R, f)$
8: **end if**
9: Generate higher-dimensional cells inductively: the $k$-simplex $[v_{a_0} v_{a_1} \cdots v_{a_k}]$ occurs iff the three lower-dimensional simplices $[v_{a_1} \cdots v_{a_k}]$, $[v_{a_0} \cdots v_{a_{k-1}}]$ and $[v_{a_0} v_{a_k}]$ all occur.

---

where $m$ is the number of simplices, $\ell_i$ is the length of the $i$-th barcode, and $M_i$ is the radius corresponding to the median of $\ell_i$. Intuitively, the shorter the lifecycle, the more unstable the corresponding simplex, and the less influence it raises to the determination of a stable and prime simplicial complex. Formally, let the length of the shorter lifecycles be $\ell_{A,i}$ ($i = 1, \cdots, s$) and the others be $\ell_{B,j}$ ($j = 1, \cdots, s'$) with $s + s' = m$, then we can rewrite Eq. (2) as:

$$R^* = \frac{\sum_{i=1}^{s} \ell_{A,i} M_i + \sum_{j=1}^{s'} \ell_{B,j} M_j}{\sum_{i=1}^{s} \ell_{A,i} + \sum_{j=1}^{s'} \ell_{B,j}}$$

$$= \frac{\sum_{i=1}^{s} \ell_{A,i} M_i}{\sum_{i=1}^{s} \ell_{A,i} + \sum_{j=1}^{s'} \ell_{B,j}} + \frac{\sum_{j=1}^{s'} \ell_{B,j} M_j}{\sum_{i=1}^{s} \ell_{A,i} + \sum_{j=1}^{s'} \ell_{B,j}} \quad (3)$$

When for all the lifecycles, we have $\ell_{A,i} \ll \ell_{B,j}$, then Eq. (3) can be approximated by:

$$R^* \approx \frac{\sum_{j=1}^{s'} \ell_{B,j} M_j}{\sum_{j=1}^{s'} \ell_{B,j}} \quad (4)$$

It indicates that the primal simplicial complex is less sensitive to those simplicial complexes with the shorter lifecycles. It is also noting that we construct a prime simplicial complex per class for classification.

### 3.2   Classifying objects based on NSC

Assuming that data distribution per class is represented by a prime simplicial complex, we attempt to classify unlabelled samples by projecting the samples to the simplices of prime simplicial complexes. With this way, we can avoid projecting them to some holes and voids that may exist in the structures. The holes and voids will lead to incorrect projection and impair the classification performance.

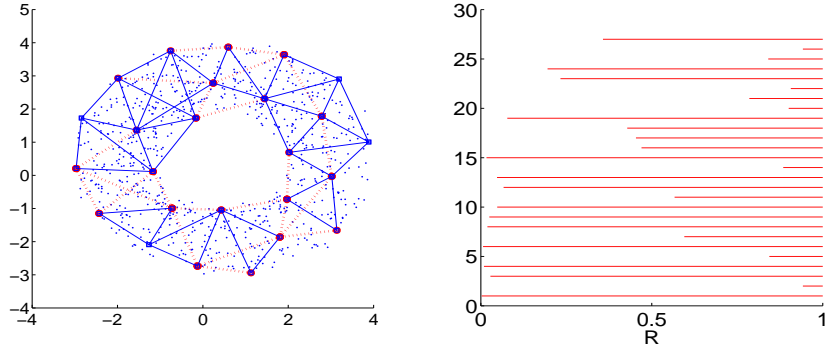

**Fig. 3.** We construct a simplicial complex (left) for a circle-shape data. In the panel, red dotted line and blue line denote 1-simplex and 2-simplex, respectively. Each barcode (right) of its simplices starts at a specific R value, and ends up at $R_{\max}$ which is used to determine when to stop the computation of barcode. In this figure, $R_{\max}$ is set to 1. Those disappeared simplices haven't been shown in the figure.

Specifically, let $\sigma_i$ $(i = 1, 2, \cdots, m)$ be a $k$-simplex with vertices $\{v_0, v_1, \cdots, v_k\}$. Then the projection position $x_p$ of sample $x$ can be defined as a linear combination of vertices in the simplex:

$$x_p = \sum_{i=0}^{k} \lambda_i v_i, \quad \text{where} \quad \sum_{i=0}^{k} \lambda_i = 1 \tag{5}$$

Here $\lambda_i$ is the weight value. Take a 2-simplex as an example. The weight is equal to

$$\lambda_i = \begin{cases} (B^T B)^{-1} B^T (x - v_i), & i = 0, 1 \\ 1 - \lambda_0 - \lambda_1, & i = 2 \end{cases} \tag{6}$$

where $B = [v_0 - v_1, v_1 - v_2]$. As for a 1-simplex, the weight is equal to:

$$\lambda_i = \begin{cases} \frac{(x - v_1)^T (v_1 - v_0)}{(v_1 - v_0)^T (v_1 - v_0)} & i = 0 \\ 1 - \lambda_0 & i = 1 \end{cases} \tag{7}$$

If the projection index $0 \leq \lambda_i \leq 1$, the projection position locates inside the face. Otherwise, it locates outside the face. For $\lambda_i > 1$ or $\lambda_i < 0$, on one hand, it can lead to an incorrect projection for distant points. On the other hand, it provides a "forward" and "backward" interpolation along a face when the number of training sample is small. To make a compromise between preventing that data are incorrectly projected outside the face and preserving the extrapolation ability of topological structure, we introduce a parameter $\gamma$ to compute the projection position and corresponding projection distance as follows:

$$x_p = \begin{cases} v_i + (1+\gamma)(v_j - v_i), & \text{if } \lambda_i \geqslant 1+\gamma \\ v_i - \gamma(v_j - v_i), & \text{if } \lambda_i \leqslant -\gamma \end{cases} \quad (8)$$

where $v_i, v_j$ denote two different vertices of a simplex. Then the distance between a sample $x$ and a simplicial complex of the $c$-th class is:

$$d_{NSC}(x|SC_c) = \min_\ell (x - x_p^{\ell,c})^T \mathcal{A}(x - x_p^{\ell,c}),$$
$$\ell = 1, \cdots, m; c = 1, \cdots, C \quad (9)$$

where $m$ denotes the number of simplices in the complex, $C$ is the number of classes, and $\mathcal{A}$ is a non-negative matrix [4]. Finally, we classify sample to a class that has the nearest simplicial complex distance to the sample:

$$\mathcal{C}(x) = \arg\min_c d_{NSC}(x|SC_c) \quad c = 1, \cdots, C \quad (10)$$

## 4  Experiments

Experiments are performed to evaluate the performance of the NSC approach. Here, two face recognition datasets and eight UCI benchmark data sets [16] are used as listed in Tab. 1.

We also use five simulated datasets and four practical multi-view datasets. The five simulated datasets are generated from different topological structures plus random noise with variance $\rho$. They are 1) D1: two concentric circles ($\rho = 1.0$); 2) D2: two spirals ($\rho = 3.5$); 3) D3: circle-cross-circle ($\rho = 2.0$); 4) D4: four circle-cross-circle ($\rho = 2.0$) and 5) D5: Sphere-cross-sphere ($\rho = 1.5$) datasets as shown in Fig. 4. Each dataset includes 2-class, each of which has 500 training samples and 500 test samples without overlap. We use max-min sampling strategy to select 50% training samples as the landmark points [4] and the remaining samples as the witness points to construct the prime simplicial complexes. Some examples of these complexes are illustrated in Fig. 4. From the figures we can see that the NSC approaches preserve the structure representation well.

The four practical multi-view data sets used for object recognition are COIL-20 [17], COIL-100 [18], SOIL-47A and SOIL-47B [19]. The COIL-20 dataset consists of 20 objects, each of which has 72 different views that are sampled every $5^o$ around an axis passing through the object. Each object is an image with size $128 \times 128$. We subsample them to $32 \times 32$ ones. The COIL-100 dataset has

---

[4] It can be obtained by metric learning which goes beyond the scope of this paper. In our paper, we set it to be an identity matrix or inverse covariance matrix. The former one is equivalent to a Euclidean distance. Meanwhile, the latter one leads to a classical Mahanalobis distance, named NSC-M.

**Table 1.** Description of several benchmark data sets. Here "#", "Dim" denote the number of samples and means dimension, respectively. 'C' denotes the number of classes, and 'RA' denotes the ratio of the number of training samples in each dataset or the number of training samples versus that of test samples. The latter one means that training set and test set have been separated by their provider.

| Datasets | # | Dim | C | RA |
|---|---|---|---|---|
| ORL | 400 | 10304 | 40 | 0.5 |
| UMIST | 575 | 10304 | 20 | 0.5 |
| Iris | 150 | 4 | 3 | 0.5 |
| Landsat Satellite | 6335 | 36 | 2 | 0.1 |
| Image Segmentation | 2310 | 16 | 7 | 210/2100 |
| Gaussian Elena | 5000 | 8 | 2 | 0.5 |
| Breast Cancer Wisconsin | 569 | 31 | 2 | 0.5 |
| Phoneme | 5404 | 5 | 2 | 0.1 |
| Pendigits | 10992 | 17 | 10 | 7494/3498 |
| Optdigits | 5620 | 65 | 10 | 3823/1797 |

**Table 2.** Experiment I: The influence of $f$ to the classification performance on the five simulated datasets. Experiment II: The influence of $R_{\max}$ and comparison with other algorithms. The experiment results are the average of 20 repetitions. In the table, $A\pm B$ means average error rate and standard deviation (%).

| | D1 | D2 | D3 | D4 | D5 |
|---|---|---|---|---|---|
| Experiment I: $R_{\max}=0.5, \gamma=0$ | | | | | |
| NSC ($f=0$) | $4.24\pm0.68$ | $13.38\pm1.18$ | $8.09\pm1.03$ | $9.08\pm1.05$ | $3.85\pm0.56$ |
| NSC-M ($f=0$) | $4.40\pm0.85$ | $13.27\pm1.10$ | $9.52\pm0.92$ | $13.29\pm1.55$ | $7.62\pm1.04$ |
| NSC ($f=1$) | $3.81\pm0.59$ | $11.70\pm1.09$ | $\mathbf{6.01\pm0.74}$ | $6.94\pm0.82$ | $2.97\pm0.57$ |
| NSC-M ($f=1$) | $3.83\pm0.59$ | $11.66\pm1.13$ | $6.54\pm0.79$ | $8.53\pm1.00$ | $5.15\pm0.92$ |
| NSC ($f=2$) | $3.87\pm0.81$ | $10.68\pm1.05$ | $6.19\pm0.70$ | $\mathbf{6.55\pm0.72}$ | $\mathbf{2.89\pm0.73}$ |
| NSC-M ($f=2$) | $\mathbf{3.79\pm0.76}$ | $\mathbf{10.63\pm1.06}$ | $6.69\pm0.63$ | $7.79\pm1.05$ | $4.29\pm0.84$ |
| Experiment II: $f=2, \gamma=0$ | | | | | |
| NSC: $R_{\max}=0.5$ | $3.41\pm0.51$ | $11.70\pm0.62$ | $5.77\pm0.42$ | $6.58\pm0.57$ | $2.68\pm0.70$ |
| NSC-M: $R_{\max}=0.5$ | $3.43\pm0.52$ | $11.67\pm0.70$ | $6.23\pm0.49$ | $8.06\pm0.76$ | $4.36\pm1.01$ |
| NSC: $R_{\max}=1.0$ | $3.83\pm0.74$ | $10.60\pm0.76$ | $6.18\pm0.46$ | $6.77\pm0.70$ | $2.42\pm0.67$ |
| NSC-M: $R_{\max}=1.0$ | $3.78\pm0.74$ | $10.59\pm0.76$ | $6.46\pm0.52$ | $7.60\pm0.89$ | $3.42\pm0.88$ |
| 1-NN | $4.58\pm0.53$ | $13.24\pm0.80$ | $7.30\pm0.88$ | $7.88\pm0.71$ | $3.13\pm0.86$ |
| 3-NN | $4.20\pm0.63$ | $11.74\pm0.87$ | $6.65\pm0.76$ | $7.09\pm0.73$ | $2.83\pm0.77$ |
| SVM-G | $\mathbf{3.24\pm0.62}$ | $\mathbf{10.23\pm0.87}$ | $\mathbf{5.46\pm0.73}$ | $\mathbf{6.30\pm0.63}$ | $\mathbf{2.35\pm0.52}$ |

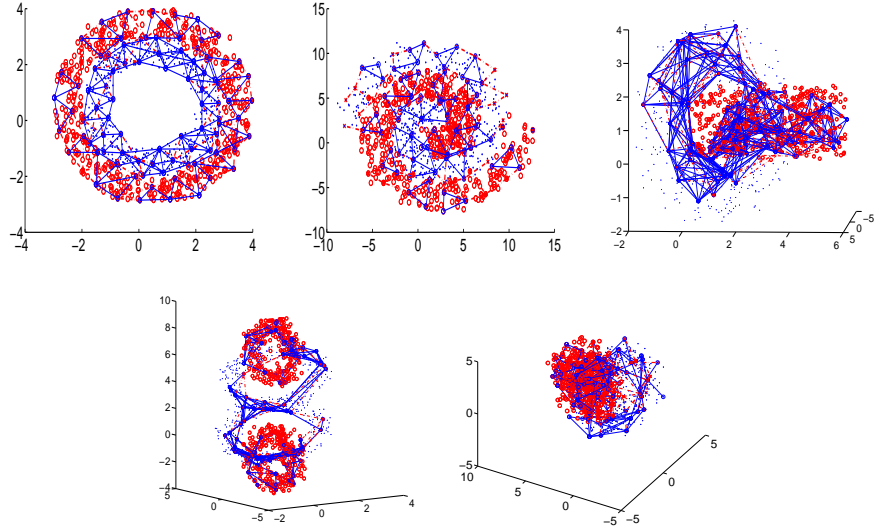

**Fig. 4.** From left to right, from top to bottom: D1 to D5 datasets. In each panel, red dotted line and blue line denote 1-simplex and 2-simplex, respectively. The test sets are generated based on the same distribution. Note that the fifth dataset cannot be shown correctly in the three-dimensional space since two spheres which cross each other can only be seen in four or higher-dimensional space.

100 objects which is collected with the same way as the COIL-20. We subsample each object image to colored $16 \times 16$ one, short for COIL-100A and gray $32 \times 32$ one, short for COIL-100B. The SOIL-47A and SOIL-47B datasets are sampled from different illuminations [19]. Each dataset consists of 47 objects, each of which has 21 different views that are sampled every $9^o$ around an axis passing through the object. Each object image is subsampled to a colored image with size $24 \times 30$. All of these images are directly served as the feature vectors. Some objects in the three practical datasets are shown in Fig. 5.

As to the two face datasets, the UMIST face dataset [20] is a multi-view one for testing the robust of our approach, and the ORL dataset [21] is also another popular benchmark one for face recognition. It is worth mentioning that our object recognition is in an instance level, i.e, all the data points in a data set belongs to the same category, and is not in the sense of the VOC (visual object classes) challenges.

For comparison, we also compare the performance of our approaches with 1-nearest neighbor (1-NN), 3-NN and SVM with Gaussian RBF kernels (SVM-G) [22]. The parameters in SVM are tuned by cross-validation. The whole training set is used by these approaches.

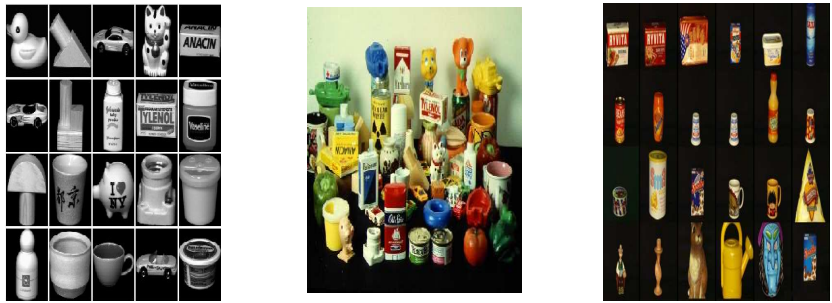

**Fig. 5.** From Left to Right: examples of COIL-20 [17], COIL-100 [18] and SOIL-47 [19] benchmark datasets.

### 4.1  Simulated datasets and parameter influences

We investigate the influence of $f$ in the five datasets. Given $f = 0, 1, 2$, $R_{\max} = 0.5$ and $\gamma = 0$, the average results of 20 repetitions are shown in Tab. 2. From the Tab. 2 we can see that the performance of the proposed approaches with $f = 2$ is better in most cases. A possible reason is that as [4] pointed out, $f = 2$ provides a clean persistent interval graph with little "noise". Therefore, it leads to a more stable structure representation. Note that in practical noisy environments, such graphs cannot be easily obtained. Meanwhile, $f = 1$ can be interpreted as arising from a family of coverings of the space X with Voronoi-like regions surrounding each landmark point. We thus set the parameter $f$ to be 1 or 2 in the subsequent experiments. Note that in "small training samples" case shown in the next subsection, we use Rips complex, which needn't the parameter $f$, for object recognition. Furthermore, we found that Mahalanobis distance is helpful to improve the performance of the proposed algorithms in some cases.

We also study the influence of the parameter $R_{\max}$ in determining the optimal value $R^*$, which is closely related to the selection of prime simplicial complex per class. We perform experiments on the five simulated datasets by selecting a group of $R_{\max}$, followed by computing the corresponding $R^*$. The results are shown in Tab. 2 and Fig. 6. From the results we can see that when $R_{\max}$ locates in an interval $[0.3, 1]$, the classification performance are better. A reason is that the radius of these simulated datasets is close to 0.5. As a result, the topological structure can be preserved well when $R_{\max}$ is selected around 0.5.

Furthermore, we compare the NSC approaches with 1-NN, 3-NN and SVM methods in the five 2-class datasets. The reported results are shown in Tab. 2. It can be seen from Tab. 2 that in these five datasets, the NSC approaches are always better than 1-NN and 3-NN, and achieve competitive performance compared with SVM-G. It is worth noting that as illustrated in Fig. 4, our approaches also preserve reasonable structure representations to these data distributions.

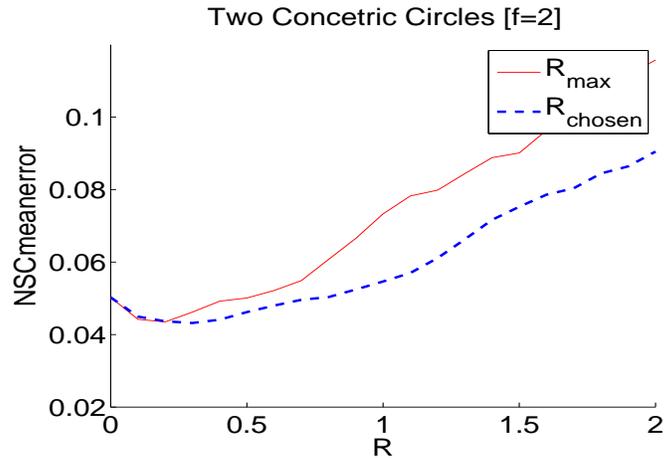

**Fig. 6.** Parameter Influence on the D1 simulated dataset.

### 4.2  Small training samples and high-dimensional datasets

We test the proposed approaches on four multi-view object recognition datasets, each of which can be regarded as generating from a circle-shape structure. We use four different views per class (i.e., $0^o, 90^o, 180^o$ and $270^o$) and eight views per class as the training set, respectively. The remaining images are used as the test set. Since the number of training samples is small, we set $\gamma$ to be 1 to enhance the extrapolation ability of NSC, and employ Rips method [4] instead for the construction of prime simplicial complexes based on the whole training set. Here $R$ is set to be 30. Note that $R$ in Rips method is different from that in Lazywitness method. The results are shown in Tab. 3. From the results we can see that compared with NN and SVM algorithms, the proposed NSC approaches achieve the best performance in 4 out of 5 datasets. In SOIL-47B dataset, the performance of NSC is slightly worse than those of SVM algorithms. It indicates that the proposed NSC approaches can work well in high-dimensional multi-view structures. Note that here we haven't reported the results of NSC-M approach since the computation of covariance matrix is ill-posed when the number of samples is less than the dimension of a data set. Furthermore, we also observe that with 8 views as the training samples, our approach obtains the competitive performance as those state-of-art algorithms using 4 views in COIL and SOIL datasets [23]. However, the latter ones utilize very effective feature extraction and image registration techniques. In contrary, our approaches achieve a good trade-off between recognition accuracy and topology preservation by only introducing additional 4 views.

**Table 3.** The error rates and standard deviations (%) of several approaches in the 12 practical datasets. Here '4V' and '8V' denote 4 and 8 views, respectively.

|  | NSC | NSC-M | 1-NN | 3-NN | SVM-G |
|---|---|---|---|---|---|
| COIL-100A (4V) | **12.84** | N/A | 16.85 | 28.40 | 14.23 |
| COIL-100A (8V) | **2.81** | N/A | 5.33 | 13.36 | 4.06 |
| COIL-100B (4V) | **24.01** | N/A | 29.50 | 43.66 | 24.23 |
| COIL-100B (8V) | **7.50** | N/A | 12.78 | 27.67 | 8.36 |
| SOIL-47A (4V) | **16.67** | N/A | 19.12 | 56.99 | 21.08 |
| SOIL-47A (8V) | **11.61** | N/A | 13.54 | 26.49 | 12.65 |
| SOIL-47B (4V) | 22.92 | N/A | 23.41 | 40.93 | **22.67** |
| SOIL-47B (8V) | 15.33 | N/A | 18.30 | 25.60 | **14.58** |
| COIL-20 (4V) | **15.00** | N/A | 16.76 | 28.24 | 17.36 |
| COIL-20 (8V) | **2.97** | N/A | 5.39 | 12.27 | 4.85 |
| ORL | $7.02 \pm 2.00$ | N/A | $8.78 \pm 2.62$ | $17.23 \pm 2.5$ | $\mathbf{6.38 \pm 2.01}$ |
| UMIST | $\mathbf{3.66 \pm 1.52}$ | N/A | $5.34 \pm 1.52$ | $11.05 \pm 2.36$ | $6.43 \pm 1.31$ |
| Iris | $5.27 \pm 2.10$ | $\mathbf{4.00 \pm 2.10}$ | $7.40 \pm 1.81$ | $5.93 \pm 2.10$ | $4.53 \pm 1.80$ |
| Landsat Satellite | $13.49 \pm 0.44$ | $20.40 \pm 0.61$ | $13.54 \pm 0.38$ | $13.58 \pm 0.55$ | $\mathbf{11.89 \pm 0.54}$ |
| Image Segmentation | 6.38 | 45.90 | 7.10 | 10.62 | **6.05** |
| Gaussian-elena | $15.24 \pm 0.58$ | $34.61 \pm 1.23$ | $20.15 \pm 0.70$ | $18.52 \pm 0.55$ | $\mathbf{9.98 \pm 0.51}$ |
| Breast Cancer Wisconsin | $\mathbf{3.77 \pm 0.95}$ | $10.63 \pm 2.03$ | $5.12 \pm 1.32$ | $4.18 \pm 0.83$ | $3.18 \pm 1.12$ |
| Phoneme | $19.91 \pm 0.61$ | $21.85 \pm 0.66$ | $16.13 \pm 0.45$ | $16.57 \pm 0.54$ | $\mathbf{15.40 \pm 0.78}$ |
| Pendigits | 2.20 | 4.20 | 2.57 | 2.43 | **1.83** |
| Optdigits | 3.06 | 3.95 | 3.45 | 3.28 | **1.56** |

### 4.3  Face recognition

We compare our approaches with others in ORL [21] and UMIST [20] face recognition datasets. In ORL dataset, the images of each subject are taken at different times with various lighting, facial expressions and facial details [21]. In UMIST dataset, the images of each subject are taken by varying angles from left profile to right profile. We employ PCA to reduce the original dimensions to 40-dimensional subspaces since empirically, the subspaces preserve most of the principal structures. Furthermore, we also employ Rips method to construct the prime simplicial complexes based on the whole training set. The results are shown in Tab. 3. It can be seen from the results that NSC approach obtains the best performance in UMIST data set and ranks 2 in ORL data set.

### 4.4  UCI datasets

Finally, we evaluate the performance of the NSC approaches in 8 UCI datasets. Different from the aforementioned datasets, these datasets are taken from remarkably different domains. The results are shown in Tab. 3. We can see from the Tab. 3 that the proposed NSC approaches achieve competitive performance in these datasets. NSC ranks 1 in 2 of 8 datasets and ranks 2 in 5 of 8 datasets. It means that although devoting to preserve structures, the proposed NSC approaches can also be applied to some general fields.

### 4.5 Discussion

Here we perform a significant analysis to the proposed NSC approaches based on the results shown in Tab. 3. With the significance level of 5%, the p-value of the paired `t-test` results for the `NSC` approaches in the 20 data sets are shown in Tab. 4. It indicates that NSC, 1-NN and SVM-G are statistically similar in these datasets.

**Table 4.** The $p$-value of the paired `t-test` results based on Tab. 3. The $p$-value in bold type indicates a rejection of the null hypothesis at the 5% significance level, which means there is significant difference between the two approaches.

|         | NSC vs 1-NN | NSC vs 3-NN | NSC vs SVM-G |
|---------|-------------|-------------|--------------|
| p-value | 0.4014      | **0.0101**  | 0.8163       |

We also want to discuss some limitations of the proposed approaches. Although our goal is to preserve the topological structure of datasets, first of all, the current persistent homology techniques can only provide some approximations to the 'truly' topological invariants, as our approaches do. It is also unclear that whether the topological structures indeed exist in the high-dimensional data sets. Secondly, the evaluation is to a certain extend unfair to our approaches since other approaches use the whole training sets to train their models, whereas due to the nature of witness complex, we have to select at most use 50% of the training sets as the landmark points to build our classification model for large-scale training samples. Thirdly, the computational complexity is higher. Given the dimension is $d$, and the number of data set is $n$, specifically, the computational complexity of Rips complexes is $O(d \cdot n^2)$, and that of witness complexes is $O(r \cdot d \cdot n^2)$, where $r$ is the ratio of the number of landmark points to $n$. Furthermore, the computational complexity of computing nearest distance from data point to the prime simplicial complexes is $O(n^2)$. When data are subject to Gaussian distribution, finally, the proposed approaches will lose their merits in recognizing objects.

## 5  Conclusion

We propose new structure-preserving NSC approaches by utilizing persistent homology technique in this paper. We refine the construction of simplicial complex by removing some simplices that are redundant to the NSC approaches. We present a new Barcode method to determine a prime simplicial complex per class for classification. We also propose a nearest projection technique by computing the distance from unlabelled samples to the prime simplicial complexes. Furthermore, we generalize the extrapolation ability of simplicial complexes with a projection constraint term. Experiments indicate that compared with several well-known algorithms, our proposed NSC approaches achieve promising performance without losing the preservation of structure representation.

In this paper, the proposed approaches does not consider how to deal with recognizing those faces in the wild. However, our goal is to design a topology-preserving classifier for object recognition and supervised learning, and the "face in the wild" problem can be avoided by employing Near-Infrared sensor to alleviate the influence of background if we attempt to employ our approach to such a scenario.

In the future, we will investigate how to employ the NSC approaches to other practical applications with more complex topological structures. Furthermore, how to construct a more suitable prime simplicial complex deserve study. Moreover, we will study the performance of the proposed NSC approaches for object recognition in a category level rather than in an instance level. Finally, we will consider to further refine the performance of the NSC approaches by utilizing metric learning methods.